\documentclass[conference]{IEEEtran}
\usepackage{amsmath,amsfonts,amssymb}
\usepackage{graphicx}
\usepackage{setspace}
\usepackage{tocloft}
\usepackage{mathtools}
\ifCLASSINFOpdf
\else
\fi

\begin{document}
\title{Superpixel Based Segmentation and Classification of Polyps in Wireless Capsule Endoscopy}
\author{\IEEEauthorblockN{Omid Haji Maghsoudi}
\IEEEauthorblockA{Department of Bioengineering, College of Engineering, Temple University, Philadelphia, PA, USA, 19122\\
Email: o.maghsoudi@temple.edu}}
\maketitle
\begin{abstract}
Wireless Capsule Endoscopy (WCE) is a relatively new technology to record the entire GI trace, in vivo. The large amounts of frames captured during an examination cause difficulties for physicians to review all these frames. The need for reducing the reviewing time using some intelligent methods has been a challenge. Polyps are considered as growing tissues on the surface of intestinal tract not inside of an organ. Most polyps are not cancerous, but if one becomes larger than a centimeter, it can turn into cancer by great chance. The WCE frames provide the early stage possibility for detection of polyps. Here, the application of simple linear iterative clustering (SLIC) superpixel for segmentation of polyps in WCE frames is evaluated. Different SLIC superpixel numbers are examined to find the the highest sensitivity for detection of polyps. The SLIC superpixel segmentation is promising to improve the results of previous studies. Finally, the superpixels were classified using a support vector machine (SVM) by extracting some texture and color features. The classification results showed a sensitivity of 91\%.
\end{abstract}

\IEEEpeerreviewmaketitle

\section{Introduction}
\begin{figure}[b!p]
\centering
\includegraphics[width=.1\textwidth]{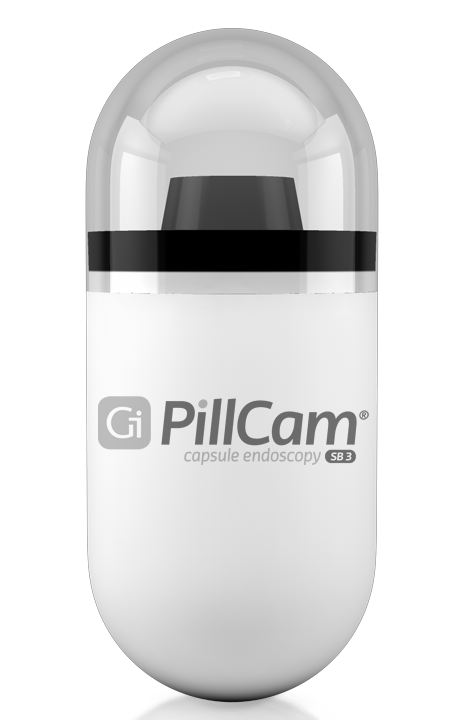}
\caption{Wireless capsule endoscopy made by Given Imaging {\cite{PillCam}}. The Figure is in color (check the DOI: 10.1109/SPMB.2017.8257027).} \label{Fig1}
\end{figure}

Wireless capsule endoscopy (WCE) is a new device which can investigate the entire gastrointestinal (GI) tract without any pain and it has been used to detect small bowel abnormalities such as polyps, tumor, bleeding, ulcers, Crohn’s disease, and Celiac. While the conventional methods are considered as painful and invasive methods. It records around 55000 frames in an examination which takes at least eight hours \cite{li2011computer}. The WCE is illustrated in Fig. \ref{Fig1}. Image processing have been used to solve problems in different applications like cell movement detection \cite{penjweini2017investigating}, animal locomotion \cite{maghsoudi17Asilomar, maghsoudi2016tracker}, or robotic \cite{pire2015stereo}. Considering a large number of frames captured, it is essential to use an intelligent software or automatic method for reviewing these frames.

There have been various investigations to help physicians for finding tumors \cite{alizadeh2014segmentation, barbosa2008detection, mahdi2017detection}, bleeding \cite{guobing2011novel, pan2011bleeding, li2009computer, maghsoudi2016detection}, Crohn's disease \cite{arguelles2014crohn, kumar2012assessment}, and different organs \cite{maghsoudi2012automatic, lee2007automatic, haji2012automatic}. 

A polyp is one of the most common diseases in the intestine among adults with a percentage of 30\% to 50\%. For people more than 50 years old, this percentage goes up to 90\% \cite{yuan2016improved}. Most polyps are not cancerous, but if one becomes larger than a centimeter, it can turn into cancer by great chance. Therefore, it is necessary to detect the polyps in the early stages.

There have been some methods tried to address this need. In the most recent study, a method was proposed to distinguish between the frames containing polyps and the one which is completely normal \cite{yuan2016improved}. Five different features, single scale-invariant feature transform (SIFT), local binary pattern (LBP), uniform LBP, complete local binary pattern (CLBP), and Histogram of Oriented Gradients (HOG), were extracted from the WCE frames. The extracted features were classified using two well-known classifiers, support vector machine and Fisher's linear discriminated analysis. The highest accuracy, specificity, and sensitivity for detection of frames containing Polyp were respectively 93\%, 94\%, and 87\%.

Another method was designed to segment the polyps in the WCE frames using two steps: first combined Log Gabor filters and SUSAN edge detector to produce potential polyp segments and then applied geometric features to outline final polyps \cite{karargyris2011detection}. 

A method was presented to detect mucosal polyps \cite{silva2014toward}. The method used edge detection techniques following by Hough transform to extract shape features and then the extracted texture features. The extracted features were provided for a Cascade Adaboost to classify the frames. The results showed a sensitivity of 91\% and a specificity of 95\%.

As mentioned, the available methods for detection of polyp region in the WCE frames are based on edge detection from grayscale image. This causes a huge disadvantage for this method as changes in contrast and signal to noise can affect the results easily. If the contrast between polyps and normal tissues cannot be noticed, it will lead to misdetection of the region. However, as the previous studies have shown, using texture and color information would be enough to find the frames containing the polyps but not to segment the region. To avoid this, we can use a consistent property of the tissue which is the color information, the RGB color space or the HSV color space. The main contribution of this study is to show the superpixel application for segmentation of polyps in the WCE frames following by classifying the superpixels. Having an accurate method for segmentation of diseases will lead to the more accurate detection of frames having these abnormalities region. We propose and evaluate a method to segment the polyp regions. Then, we extract some texture and color features and feed these features to an SVM to classify the regions.

\section{Methods}
\subsection{SLIC Superpixel Segmentation}
Segmentation has played a great role in different applciations \cite{minaee2016screen}. Superpixel contracts and groups uniform pixels in an image and it has been widely used in many computer vision for segmentation, object recognition, and tracking objects in videos. The superpixel concept was presented as defining the perceptually uniform regions using the normalized cuts (NCuts) algorithm. \cite{Li12, Levinshtein09}. Here, we use simple linear iterative clustering (SLIC) superpixel segmentation on different color images {\cite{Achanta12}}. 

The key parameter for SLIC is the size of superpixel. First, $N$ centers are defined as cluster centers. Then, to avoid keeping the center to be on the edge of an object, the center is transferred to the lowest gradient position in a $3\times3$ neighborhood. The next step is clustering, as each of the pixels is associated with the nearest cluster center based on color information. It means that two coordinate components ($x$ and $y$) depict the location of the segment and three components (for example in the RGB color space, $R$, $G$, and $B$) are derived from color channels. SLIC tries to minimize a distance (a Euclidean distance in 5D space) function is defined as follow {\cite{maghsoudi17Asilomar}}:
\begin{equation}
\label{eq:1}
D_{c} = \sqrt{(R_{j}-R_{i})^{2}+(G_{j}-G_{i})^{2}+(B_{j}-B_{i})^{2}},
\end{equation}
\begin{equation}
\label{eq:2}
D_{p} = \sqrt{(x_{j}-x_{i})^{2}+(y_{j}-y_{i})^{2}},
\end{equation}
\begin{equation}
\label{eq:3}
D = \sqrt{(\frac{D_{c}}{N_{c}})^{2}+(\frac{D_{p}}{N_{p}})^{2}}.
\end{equation}
where $N_{c}$ and $N_{p}$ are respectively maximum distances within a cluster used to normalized the color and spatial proximity. It should be said that SLIC is not letting the region grows more than twice of cluster radius; therefore, SLIC size plays an important role on how the segmentation is performed.
\begin{figure}[t!p]
\centering
\includegraphics[width=0.45\textwidth]{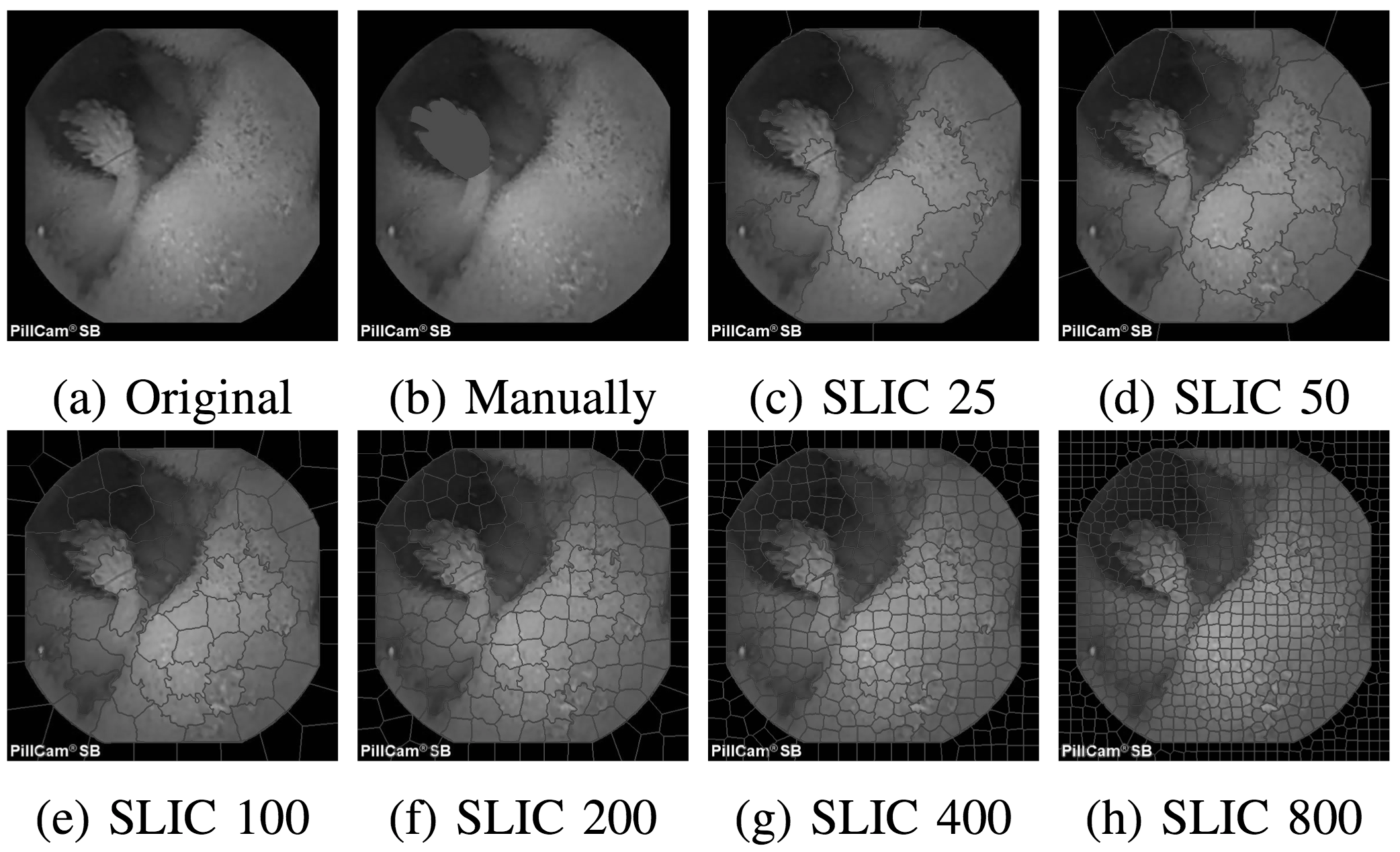}
\caption{SLIC superpixel segmented regions for a sample frame. A, b, c, d, e, f, g, and h are respectively original frame, manual outlined region for polyp, SLIC by 25, 50, 100, 200, 400, and 800. The Figure is in color (check the DOI: 10.1109/SPMB.2017.8257027).} \label{Fig2}
\end{figure}
\subsection{Features and Classifier}
\begin{figure}[b!p]
\centering
\includegraphics[width=0.47\textwidth]{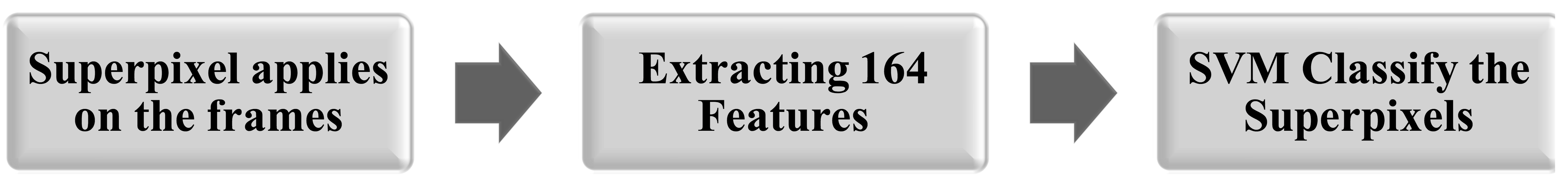}
\caption{The flowchart sumerizing the steps. The Figure is in color (check the DOI: 10.1109/SPMB.2017.8257027).} \label{Fig5}
\end{figure}

The following features were extracted from the generated superpixels: 
\begin{itemize}
  \item Grayscale and rotation invariant local binary pattern (GRILBP) was calculated for blocks of $3 \times 3$ pixels \cite{ojala2000gray}. The calculated GRILBP was mapped into a histogram having 32 bins providing 32 features for a classifier.
  \item A total of 18 Haralick features using gray level co-occurrence matrix (GLCM) were extracted from each of the generated GRILBP results, grayscale image, red, green, blue, and hue channels \cite{maghsoudi2016detection}. These features are included autocorrelation, cluster prominence, energy, cluster shade, dissimilarity, contrast, entropy, homogeneity, maximum probability, correlation, the sum of variance squares, sum average, sum variance, sum entropy, difference variance, difference entropy, information measure of correlation and inverse difference momentum. Therefore, 108 features were extracted using the Haralick features.
    \item In addition to the Haralick features, statistical features including Mean, Variance, Skewness, and Kurtosis were extracted from the same six images mentioned above.
\end{itemize}
In total, 164 features were extracted to feed an SVM to classify the superpixels to two classes: normal and polyp. The SVM used least square as a statistic measure and Gaussian radial basis function as training kernel. The steps are summarized in Fig. \ref{Fig5}.

\section{Results}
43 frames containing polyps from 5 patients were used in this study. Each of the frames was captured by an 8-bit color depth and the resolution $576\times576$. The polyp region in the frames was manually outlined.

To evaluate the SLIC superpixel for detection of polyps in the WCE frames, SLIC was applied on the frame by 6 different sizes: 25, 50, 100, 200, 400, and 800. The maximum size of SLIC is coming from the fact the minimum polyp size was 150 pixels. As a number of pixels after creating superpixels by SLIC can be up to twice of the initial size \cite{Achanta12}, the following calculation helps us to find the maximum required superpixel size for detection of a region by 150 pixels:
\begin{equation}
Maximum\ required\ SLIC\ size = (576*576)/(2*150)
\end{equation}
Fig. \ref{Fig2} shows how the SLIC superpixel size can affect on segmentation of regions. 

\begin{figure}[t!p]
\centering
\includegraphics[width=0.48\textwidth]{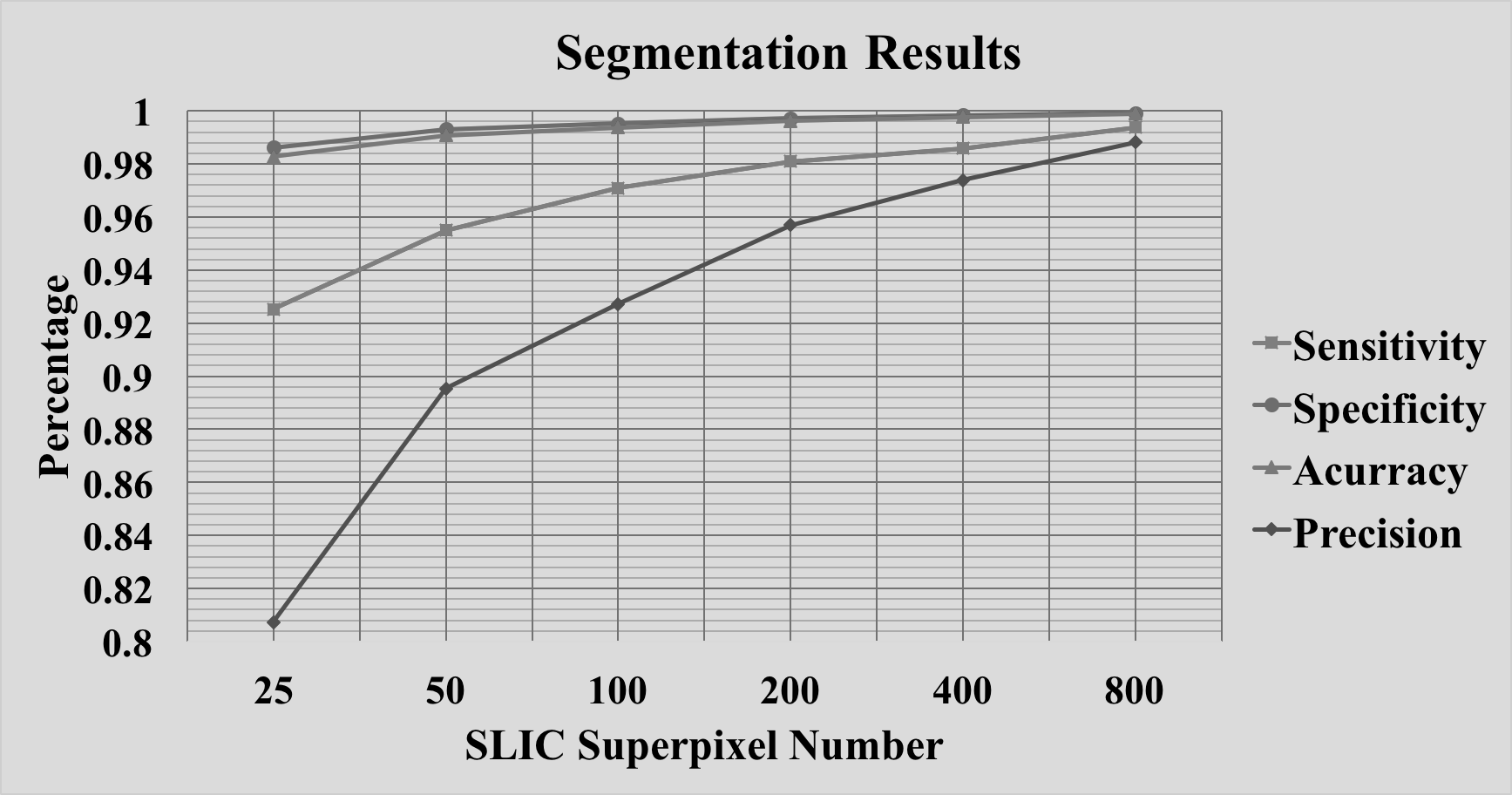}
\caption{Four measurements are graphed for different SLIC superpixel numbers. The sensitivity, specificity, accuracy, and precision are in red, blue, green, and purple, respectively. The Figure is in color (check the DOI: 10.1109/SPMB.2017.8257027).} \label{Fig3}
\end{figure}

The segmented region using the proposed method was compared with a manual outlined region of the frame. Sensitivity, specificity, accuracy, and precision are measured to do this comparison \cite{maghsoudi2012automatic}. The results are graphed in Fig. \ref{Fig2} and quantified in Fig. \ref{Fig3}. 

By evaluating the segmentation of polyps using SLIC method, it was noticed that superpixels could segment the polyp region very accurate; therefore, the remained question was how accurate the superpixels can be classified into the polyps and normal region. Therefore, the 164 features were extracted and normalized between 0 and 1. The normalized features were fed to the SVM. Three patients were randomly selected which created 27 frames for the training of SVM. The other 16 frames from two patients were used to test the classification. The outcome of SVM was evaluated based on two aspects: frame and pixel-based detection. Pixel-based detection evaluation reported the results considering how accurate the superpixels were classified as polyps and normal and the frame based detection considered a frame as polyp if there was just one superpixel in that frame classified as a polyp.

\begin{figure*}[t!p]
\centering
\includegraphics[width=0.55\textwidth]{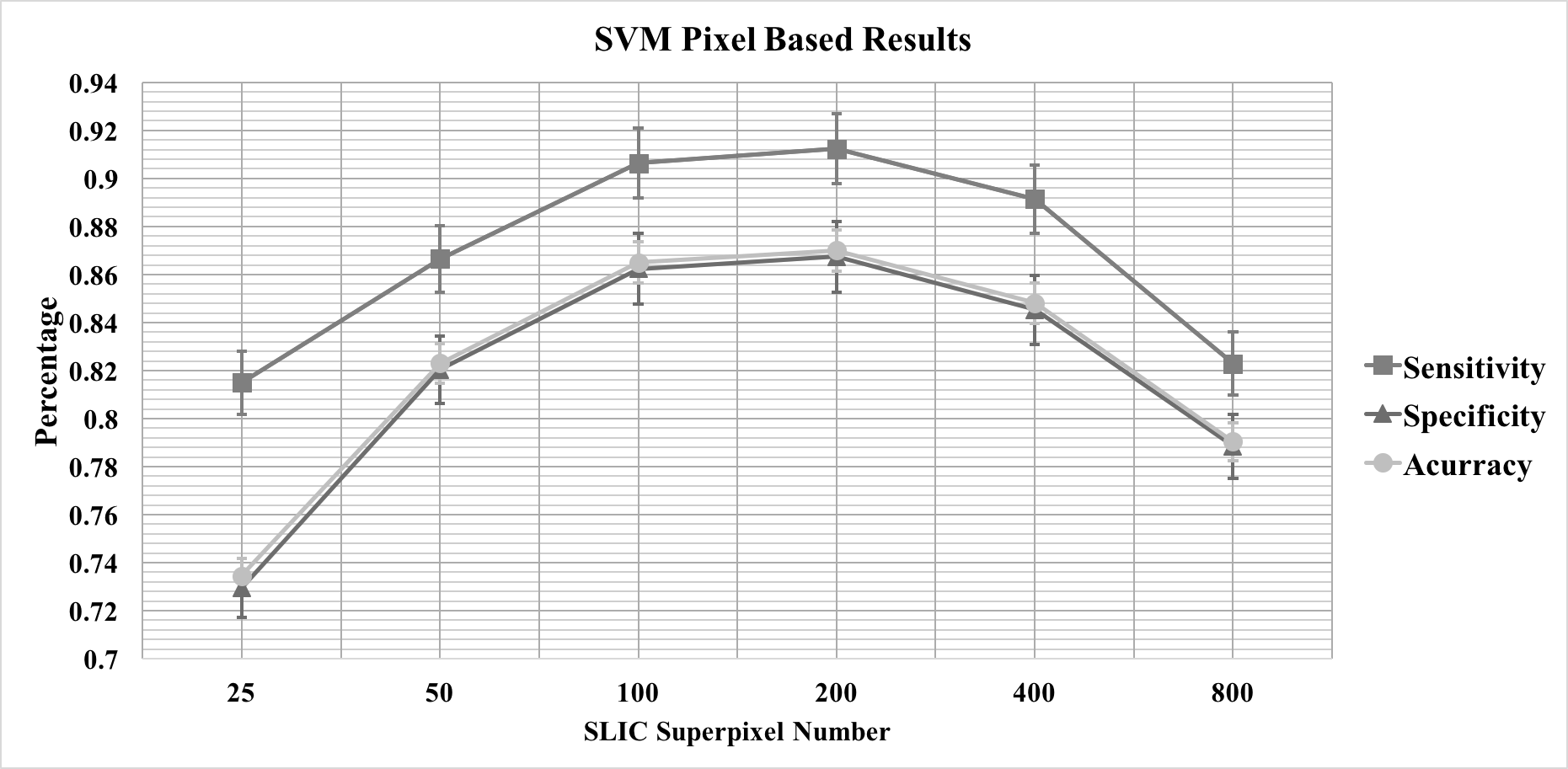}
\caption{The pixel-based results after using SVM. The sensitivity, specificity, and accuracy are in red, blue, green, and purple, respectively. The error bars show the measures standard deviation between the testing frames. The Figure is in color (check the DOI: 10.1109/SPMB.2017.8257027).} \label{Fig4}
\end{figure*}

The pixel-based evaluation was performed just for the frames having polyps (16 frames assigned for SVM testing) and the results are illustrated in Fig. \ref{Fig4}. As seen in Fig. \ref{Fig4}, the measures reached to peak after 100 superpixels and dropped dramatically after 200 superpixels. A possible reason can be the fact that when the number of superpixels is getting higher, the texture information might reduce. This limit has been seen after 200 superpixels in this study; therefore, the optimum number of superpixel can be considered as 100 as it reached the peak and generating 100 superpixels is faster than 200. Therefore, we used 100 superpixels for the frame based evaluation. In addition, to have balanced classification frame numbers, 21 frames containing normal regions were added to our database and the results are shown in TABLE \ref{tab1}. 

\begin{table}[t!p]
\caption{The frame based results for 16 polyp and 21 normal frames (the confusion matrix).} 
\label{tab1}
\begin{center}    
\begin{tabular}{ | c | c | c |}
    \hline
     & Polyp Frames & Normal Frames\\ \hline
    Predicted as Polyp & 15 & 2 \\ \hline
    Predicted as Normal & 1 & 19 \\ \hline
\end{tabular}
\end{center}
\end{table}

\section{Conclusion}
The possibility of polyps segmentation in the WCE frames was evaluated using SLIC superpixel method. The SLIC superpixel was applied on the frame with different sizes to study the effect of SLIC size for the segmentation. By increasing the size of SLIC superpixel, the four measures increased as expected because it is getting closer to pixel segmentation. 

It can be seen in Fig. \ref{Fig3} that the four measures, sensitivity, specificity, accuracy, and precision, were higher than 98\%. This gave us the possibility to detect the polyp regions very accurate; therefore, texture and color features were extracted, and then, an SVM classifier was utilized to differentiate between polyp and normal regions. 

The outcome of the classifier was evaluated in two ways: frame and pixel-based detection. The pixel-based method results are illustrated in Fig. \ref{Fig4}; the results show that the sensitivity, specificity, and accuracy were increasing then decreasing by growing the number of superpixels. These measures remained almost constant after 100 superpixels and dropped after 200. Therefore, 100 superpixels were considered as an optimum number of superpixels and the frame based aspect was studied for 100 superpixels; the frame based detection results showed a sensitivity of 93.75\%; the detail can be found in TABLE \ref{tab1}. 

Compared to the available frame based methods for detection of polyps, the sensitivity was reported 93\% \cite{yuan2016improved}, our results indicated 0.75\% higher in sensitivity. Our method also met the same detection rate for pixel-based detection. This results could have been predicted because the superpixel segmentation was so accurate as seen in Fig. \ref{Fig3}. The accurate segmentation following by feature extraction and a good classifier lead to high accurate result for classification seen in Fig \ref{Fig4}.

We will try to merge the superpixels with more efficient methods that might improve the results we presented here. That being said that the geometry features were not used in our study; by merging the superpixels, we might be able to use these types of features which can improve the detection rate. Deep learning \cite{minaee2016palmprint, minaee2016experimental} can be used to improve the classification and feature extraction process.

\section{Acknowledgements}
We have been grateful for getting help from Dr. Hossein Asl Soleimani.
\bibliographystyle{IEEEtran} 
\bibliography{report}

\end{document}